\def\year{2021}\relax
\documentclass[letterpaper]{article} 
\usepackage{aaai21}  
\usepackage{times}  
\usepackage{helvet} 
\usepackage{courier}  
\usepackage[hyphens]{url}  
\usepackage{graphicx} 
\usepackage{natbib}  
\usepackage{caption} 
\usepackage{latexsym}
\usepackage{booktabs}
\usepackage{multirow}
\usepackage{graphicx}
\usepackage{amsmath}
\usepackage{amsthm}
\usepackage{booktabs}
\usepackage{algorithm}
\usepackage{algorithmic}
\usepackage{verbatim}

\usepackage{microtype}

\title{Multiple Structural Priors Guided Self Attention Network for Language Understanding}
\author{
    Le Qi, \textsuperscript{\rm 1}
    Yu Zhang, \textsuperscript{\rm 1}
    Qingyu Yin, \textsuperscript{\rm 1}
    Ting Liu \textsuperscript{\rm 1} \\
}

\affiliations{
    \textsuperscript{\rm 1} Research Center for Social Computing and Information Retrieval, Harbin Institute of Technology, Harbin, China\\
    \{lqi,zhangyu,qyyin,tliu\}@ir.hit.edu.cn

}

\begin{document}
\maketitle
\begin{abstract}
Self attention networks (SANs) have been widely utilized in recent NLP studies. Unlike CNNs or RNNs, standard SANs are usually position-independent, and thus are incapable of capturing the structural priors between sequences of words. Existing studies commonly apply one single mask strategy on SANs for incorporating structural priors while failing at modeling more abundant structural information of texts. In this paper, we aim at introducing multiple types of structural priors into SAN models, proposing the Multiple Structural Priors Guided Self Attention Network (MS-SAN) that transforms different structural priors into different attention heads by using a novel multi-mask based multi-head attention mechanism. In particular, we integrate two categories of structural priors, including the sequential order and the relative position of words. For the purpose of capturing the latent hierarchical structure of the texts, we extract these information not only from the word contexts but also from the dependency syntax trees. Experimental results on two tasks show that MS-SAN achieves significant improvements against other strong baselines.
\end{abstract}


\section{Introduction}
Self attention networks (SANs) have been widely studied on many natural language processing (NLP) tasks, such as machine translation \cite{vaswani2017attention}, language modeling \cite{devlin2019bert} and natural language inference \cite{guo2019gaussian}.
It is well accepted that SANs can leverage both the local and long-term dependencies through the attention mechanism, and are highly parallelizable thanks to their position-independent modeling method.

However, such position-independent models are incapable of explicitly capturing the boundaries between sequences of words, thus overlook the structure information that has been proven to be robust inductive biases for modeling texts \cite{guo2019star}. Unlike RNNs that model sequential structure information of words by using memory cells, or CNNs that focus on learning local structure dependency of words via convolution kernels, SANs learn flexible structural information in an indirect way almost from scratch. One way to integrate structural information into SAN models is via pre-training, such as BERT \cite{devlin2019bert}, which learns to represent sentences by using unsupervised learning tasks on the large-scale corpus. Recent studies \cite{hewitt2019structural} have shown the ability of pre-training models on capturing structure information of sentences.

Another method to deal with structural information is introducing structure priors into SANs by mask strategies. \citeauthor{shen2018disan} \shortcite{shen2018disan} proposed the directional self-attention mechanism, which employs two SANs with the forward and backward masks respectively to encode temporal order information. \citeauthor{guo2019gaussian} \shortcite{guo2019gaussian} introduced the Gaussian prior to the transformers for capturing local compositionality of words. Admittedly, structure priors can strengthen the model's capability of modeling sentences and meanwhile assist in capturing proper dependencies. With the help of these learned structure priors, SANs can model sentences accurately even in resource-constrained conditions.
\begin{figure*}[!t]
	\centering
	\includegraphics[scale=0.36]{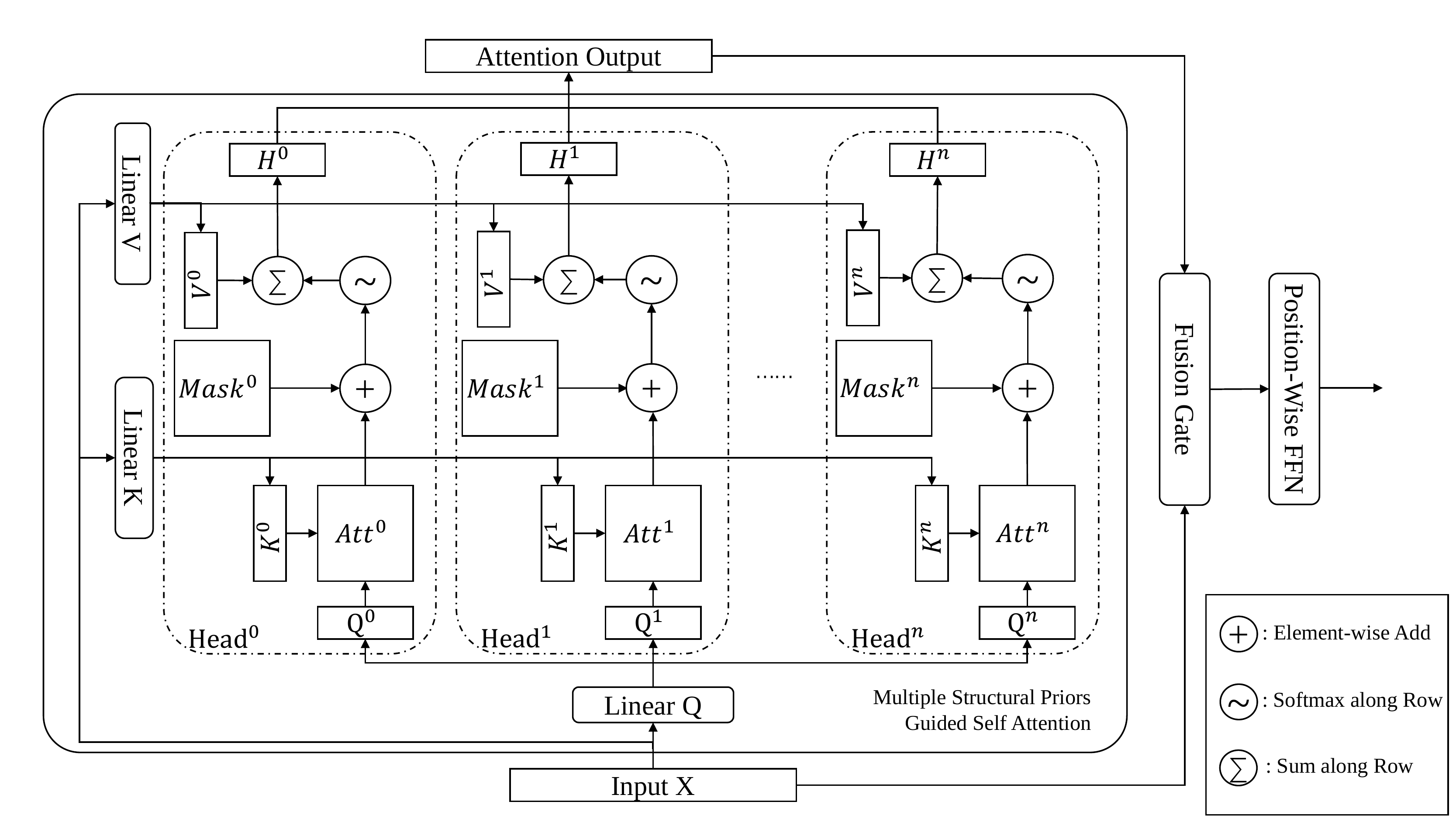}
	\caption{Architecture of the MS-SAN encoder.}
	\label{fig:1}
\end{figure*}

Though these models get success on many NLP tasks, these studies commonly focus on integrating one single type of structure priors into SANs, thus fail at making full use of multi-head attentions. One straightforward advantage of using the multi-head attentions lies in the fact that different heads convey different views of texts \cite{vaswani2017attention}. In other words, multi-head attentions enable the model to capture the information of texts at multiple aspects, which in return brings thorough views when modeling the texts. 
Besides, it is well accepted that one type of structural prior can only reveal part of the structural information from one single perspective. A variety of types of structural priors are needed in order to gain complete structural information of texts. This can be achieved by introducing different structural priors into different parts of attention heads, where different structural priors can complement each other, guiding the SAN models to learn proper dependencies between words. Therefore, to gain a better representation of the texts, a desirable solution should make full use of the multi-head attention mechanism and utilize multiple types of structural priors.

To better alleviate the aforementioned problems, in this paper, we propose a lightweight self attention network, i.e., the Multiple Structural Priors Guided Self Attention Network (MS-SAN). The novel idea behind our model lies in the usage of the multi-mask based multi-head attention (MM-MH Attention), which helps our model to better capture different types of dependencies between texts. Thanks to the MM-MH Attention mechanism, our model can capture multiple structural priors, which in return brings benefits in modeling sentences.

Especially, the structural priors we employed come from two categories: the sequential order and the relative position of words. Since the standard SANs are incapable of distinguishing the order between words, we apply the direction mask \cite{shen2018disan} directly to each attention head. Motivated by the Bidirectional RNNs \cite{schuster1997bidirectional}, we split the attention heads into two parts. For a given word, we apply the forward mask to the first half of attention heads, which allows it to attend on only the previous words when modeling the reference word. Accordingly, the backward mask is applied to the rest of the attention heads.

Since the direction masks take no consideration of the difference between long-distance words and nearby words, we employ the second category of structural prior as a complement, which could be measured by the distance between pair of words. We integrate two types of distance masks into different attention heads. The first one we utilized is the word distance mask, which describes the physical distance between each pair of words. Besides, for the purpose of capturing the latent hierarchical structure of sentences, we integrate another kind of distance information, i.e., dependency distance that is defined as the distance between each pair of words on a dependency syntax tree. The word distance mask helps our model to focus on the local words and the dependency distance mask enables our model to capture the hierarchical relationships between words. Consequently, they provide our model the ability of capturing the local and non-local dependency of words properly.

To illustrate the effectiveness of our model, we conduct experiments on two NLP tasks: natural language inference and sentiment classification. Experimental results show that MS-SAN outperforms other baselines and achieves a competitive performance comparing with the state-of-the-art models. 

Our contributions are listed as follows:
\begin{itemize}
\item In order to make full use of the multi-head attention and integrate multiple priors into a unified model, we propose a multi-mask based multi-head attention, assigning individual inductive bias to each attention head to guide the model learning more precise dependencies.

\item We utilize two categories of structure priors in our model, including the sequential order and the relative position of words. They together benefit our model on modeling sentence structures and revealing relationships between texts from multiple perspectives.

\item We empirically illustrate the effect of our model on benchmark datasets.
\end{itemize}

\section{Methodology}
Our multiple structural priors guided self attention network (MS-SAN) is composed of a stack of encoders and a pooling layer. As shown in Figure \ref{fig:1}, there are three main components in the encoder: multiple structural priors guided self attention layer, fusion gate, and position-wise feed-forward network (FFN). We describe the three components and the pooling layer in detail in the following subsections.

\begin{figure*}[!t]
	\centering
	\includegraphics[scale=0.33]{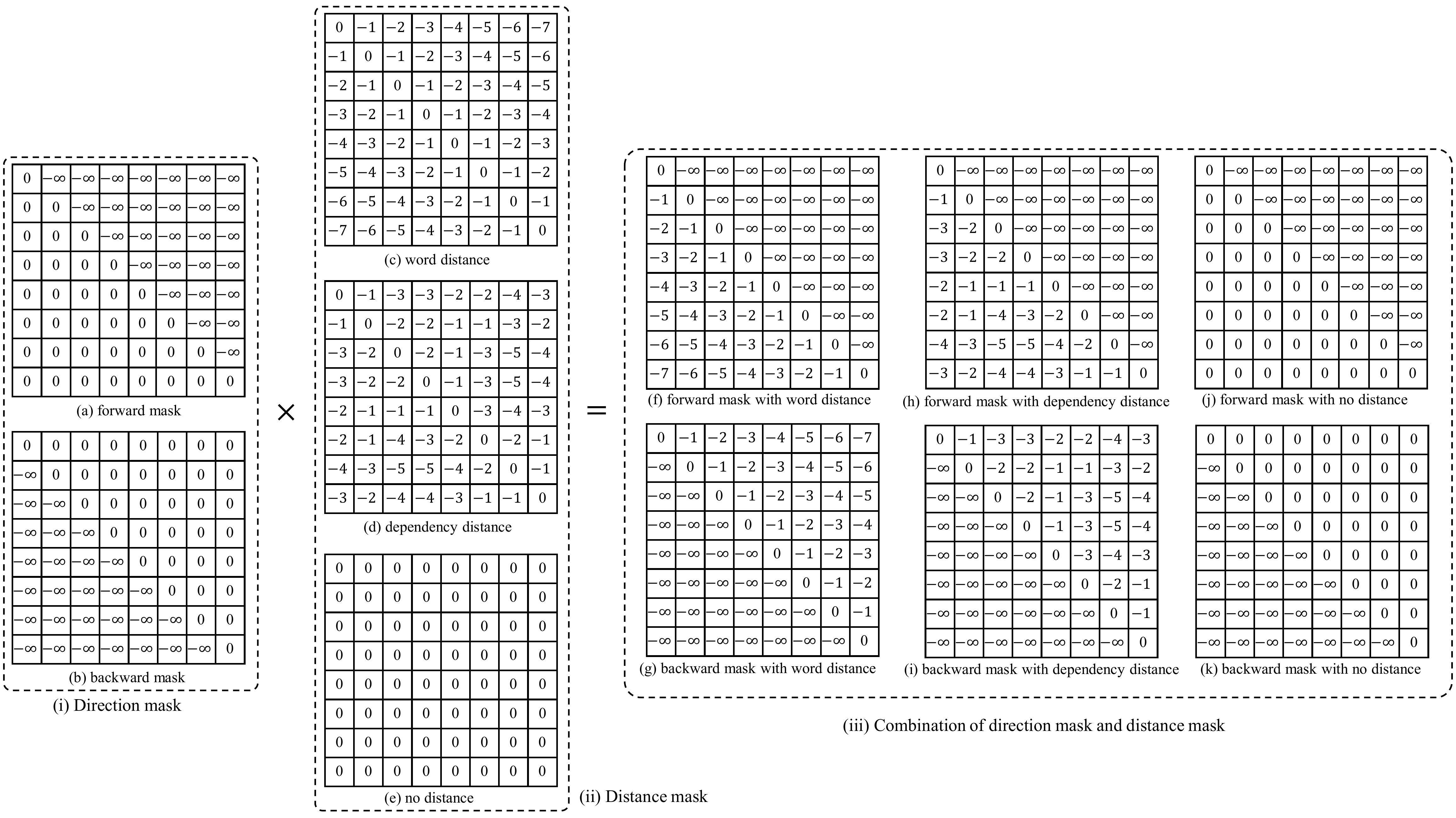}
	\caption{An example of the structural prior based masks, including direction masks, distance masks and the combination of these two kinds of masks.}
	\label{fig:2}
\end{figure*}

\subsection{Multiple Structural Priors Guided Self Attention Layer}
The multiple structural priors guided self attention layer is the core component of MS-SAN. In this subsection, we first introduce the multi-mask based multi-head attention mechanism, which is utilized to merge different structural priors into different attention heads. After that, we describe in detail about the structural priors we select and the method employed to combine these priors.

\subsubsection{Multi-Mask Based Multi-Head Attention}
Our multi-mask based multi-head attention (MM-MH Attention) is a variation of the multi-head attention proposed by \cite{vaswani2017attention}, which is combined by a set of scaled dot-product attentions. In MM-MH attention, we apply the positional masks to the scaled dot-product attentions through an addition operation, expressed as:
\begin{equation}
Att(Q,K,V,M)=\mathrm{softmax}(\dfrac{QK^{T}}{\sqrt{d_{k}}}+M)V \label{1}
\end{equation}
where $Q$, $K$, $V$ are attention matrices composed of a set of queries, keys and values, respectively, and $M$ is the mask matrix generated according to specific priors. Then, we calculate the MM-MH attention by using the following equation:
\begin{equation}
\mathrm{MM}\text{-}\mathrm{MH}(Q,K,V,M)=\mathrm{Cat}(H_{1},...,H_{n})W^{O} \label{2}
\end{equation}
where $H_{h}=Att(QW_{h}^{Q},KW_{h}^{K},VW_{h}^{V},M_{h})$, with $n$ as the number of heads, $W_{h}^{Q}, W_{h}^{K}, W_{h}^{V} \in R^{d_e \times d_e/n}$, and $W^{O}_{h} \in R^{d_e \times d_e}$. $Q=K=V \in R^{l \times d_e}$ are the embedding matrices created from a sentence with length $l$ and $M_h \in R^{l \times l}$ is the positional mask matrix for head $h$.

The MM-MH attention first projects $Q,K,V$ into $n$ sub-spaces, then creates $n$ mask matrices for each head with different priors, and generate masked attention following equation \eqref{1}, respectively. Finally, the $n$ attention results are concatenated together before the projection.

\begin{figure}[!t]
	\centering
	\includegraphics[scale=0.4]{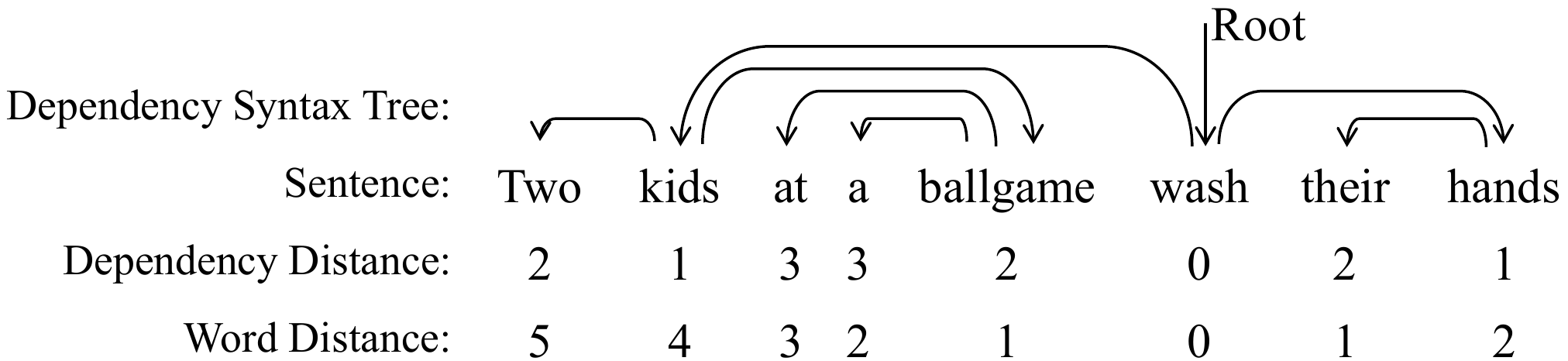}
	\caption{Comparison between word distance and dependency distance.}
	\label{fig:3}
\end{figure}
\subsubsection{Structural Priors}
Inspired by the traditional CNNs, RNNs and tree-based neural network models, we divide the structural priors into two categories: the sequential order; and the relative position of words, which could be measured by the distance between pair of words. In particular, we introduce the direction mask for the former and the word distance mask along with the dependency distance for the latter. 

\paragraph{Direction Mask}
The standard SANs can capture dependencies from the whole sequence for the reference word, ignoring the location of its contextual words. It is hard for the standard SANs to identify which word appears before or after the reference word. However, previous studies have proved these information are beneficial on modeling sentences. Therefore, following the DiSAN \cite{shen2018disan}, we apply the direction mask to the original attention distribution. As shown in Figure \ref{fig:2}(i), the forward mask and the backward mask matrices are calculated as:
\begin{flalign}
M_{ij}^{f} = \begin{cases}
0, & \text{ if } i \leq  j\\ 
-\infty,  & \text{ if } i> j
\end{cases} \\
M_{ij}^{b} = \begin{cases}
0, & \text{ if } i \geq  j\\ 
-\infty,  & \text{ if } i < j
\end{cases}
\end{flalign}
where $M_{i,j}$ is the element in the i-th row and j-th column of the mask matrix $M$.

The generated attention will ignore all words appearing after the reference word if using forward masks, and do the reverse with backward masks.
\paragraph{Word Distance Mask}
The word distance describes the relative positions of words and is commonly utilized for modeling the local compositionality. The word distance masks make our model capable to distinguish between long-distance words and nearby words, which is commonly ignored by using only direction masks. Through the word distance mask, our model can attend more on the neighboring words and thus capture more local dependencies. As shown in Figure \ref{fig:2}(c), we calculate the word distance mask $(i,j)$ by using the reverse of the relative distance between word $i$ and word $j$:
\begin{equation}
	M_{ij}^{w}= -|i-j|
\end{equation}

\paragraph{Dependency Distance Mask}
The dependency distance describes the relative positions of words on the dependency syntax tree and can support modeling long-distance dependencies and the latent hierarchical structures. According to the assumption that the closer the words on the dependency syntax tree is, the stronger the semantic relationship between them will be, the dependency distance is calculated as the reverse of the relative distance on the dependency syntax tree.
 
Given the sentence ``Two kids at a ballgame wash their hand'' and the reference word ``wash'', the dependency syntax tree and the two distances are shown in the Figure \ref{fig:3} and the dependency distance mask matrix is shown in Figure \ref{fig:2}(d). Comparing with the word distance, ``wash'' and ``kids'' are more closer according to the dependency distance. However, there are 3 words at the same distance from ``wash'' on the dependency syntax tree. Therefore, dependency distance focuses more on the non-local compositionality but cannot distinguish some words due to their same distance from the reference word. Therefore, the two distances can complement each other to better model sentences.

\paragraph{Combination of Different Masks \label{comb}}
The final mask matrix of each attention head is the combination of the direction mask and the distance mask. For head $i$, its mask matrix $M_i$ is created as:
\begin{equation}
M_i = M_{dir}^{i} + \alpha M_{dis}^{i}
\end{equation}
where $\alpha$ is a hyperparameter. $M_{dir}^{i}$ is selected from $M^{f}$ and $M^{b}$. $M_{dis}^{i}$ is selected from $M^{w}$, $M^{dp}$ and $M^{n}$, where $M^{dp}$ is the dependency distance mask and $M^{n}$ means no distance mask is applied. 

More specifically, we employ $2h$ attention heads, applying forward masks to the first $h$ heads and backward masks to the rest. Moreover, the distance masks applied for each part of attention heads are in the same order, which means that the same type of distance mask is applied on the $i$-th and $i\text{+}h$-th attention head, where $i=\{1, 2, ..., h\}$. Figure \ref{fig:2}(iii) shows an example. Given $6$ attention heads, we apply the forward mask on the first $3$ heads and apply the backward mask on the rest ones. Among these heads, the first and fourth heads are applied by the word distance mask; the second and fifth heads are applied by the dependency distance mask.

\subsection{Fusion Gate}
Instead of the standard residual connection, we apply a more flexible fusion method. Given the original input $I$ and the attention output $O$, the fusion result is calculated as:
\begin{flalign}
	&\hat{I} = W_{I}I,\quad  \hat{O} = W_{O}O \\
	&f = sigmoid(W_{1}\hat{I}+W_{2}\hat{O}+b) \\
	&Gate(I,O) = f\odot \hat{I} + (1-f)\odot \hat{O}
\end{flalign}
where $W_{I}, W_{O}, W_{1},W_{2} \in R^{d_e \times d_e}$ and $b\in R^{d_e}$.
\subsection{Position-Wise FFN}
After the fusion gate, we follow the standard SANs and employ the position-wise FFN, which consists of two linear transformations with an activation function. Note that the position-wise FFN is combined with the residual connection and the layer normalization.
\begin{equation}
\mathrm{FFN}(x) = \delta(0, xW_1+b_1)W_2+b_2
\end{equation}
where $\delta$ is the activation function, $W_1 \in R^{d_e \times d_{h}}$, $W_2 \in R^{d_{h} \times d_e}$,  $b_1 \in R^{d_{h}}$, and $b_2 \in R^{d_e}$. 

\subsection{Pooling Layer}
Finally, we generate the ultimate representation of the input sentence through the pooling layer after the MS-SAN encoder. We employ an attentive pooling and a max pooling operation and concatenate the results together. Same as the DiSAN \cite{shen2018disan}, the attentive pooling treats each element in the vector as a feature, and apply the attention operation on them, expressed as:
\begin{equation}
\mathrm{Pool_{att}}(U) = softmax(\mathrm{FFN}(U))\odot U
\end{equation}
where FFN is shown in the equation (10) and the softmax is performed on the row dimension.

\begin{table*}[]
\centering
\resizebox{\textwidth}{!}{
\begin{tabular}{@{}llcccccc@{}}
\toprule
& Model & Dim & $|\theta|$ & SNLI & MNLI (m/mm) & SST-2 & SST-5 \\
\midrule
\multirow{5}{*}{LSTM-based} & LSTM (SNLI) \cite{bowman2016fast}           & 300D  & 3.0m   & 80.6 & -      & -    & -          \\
                            & BiLSTM (SST)\cite{li2015tree}                 & -     & -       & - & -        & -    & 49.8       \\
                            & TreeLSTM* \cite{choi2018learning}             & 600D  & 10m      & 86.0 & -    & -    & -          \\
                            & TreeLSTM* \cite{tai2015improved}              & 150D  & 316K  & -    & -       & -    & 51.0       \\
                            & HBMP \cite{talman2019sentence}                & 600D  & 22m    & 86.6  & 73.7/\underline{73.0}     & -    & -          \\
\midrule
\multirow{2}{*}{CNN based}  & Tree-CNN* \cite{mou2016natural}               & 300D  & 3.5m  & 82.1  & -      & -    & -          \\
                            & DSA \cite{yoon2018dynamic}                    & 600D  & 2.1m   & \underline{86.8}  & -     & \underline{88.5}    & 50.6       \\
\midrule
\multirow{6}{*}{SAN based}  & Transformer \cite{guo2019star}                & 300D  & -      & 82.2  & -     & -    & 50.4       \\
                            & DiSAN \cite{shen2018disan}                    & 600D  & 2.4m & 85.6  & 70.9/71.4       & -    & 51.7       \\
                            & Star-Transformer \cite{guo2019star}           & 300D  & -      & 86.0  & -     & -    & \underline{52.9}       \\
                            & PSAN* \cite{wu2018phrase}                     & 300D  & 2.0m  & 86.1  & -      & -    & -          \\
                            & Reinforced self-att \cite{shen2018reinforced} & 300D  & 3.1m    & 86.3 & -     & -    & -          \\
                            & Distance-based SAN \cite{im2017distance}      & 1200D & 4.7m    & 86.3 & \underline{74.1}/72.9     & -    & -          \\
\midrule
\multirow{2}{*}{Bert based} &BERT-base \cite{devlin2019bert}                 & 768D  & 112m  & 85.7 & 75.6/75.3    & 93.5 & 56.1      \\
                            &BERT-large \cite{devlin2019bert}               & 1024D & 339m &86.3 & 76.7/76.4    & 94.9 & 57.5      \\
\midrule
\multirow{3}{*}{Our model}  & MS-SAN*                                       & 600D  & 1.8m  & 87.0 & 73.9/73.1  & 88.6 & 53.5       \\
           & MS-SAN+$\text{BERT}_{\text{base}}$*           & 1576D & 120m  & 87.9 & 77.9/78.3  & 94.1 & 58.2       \\
           & MS-SAN+$\text{BERT}_{\text{large}}$*           & 2048D & 354m  & \textbf{88.2} & \textbf{78.7/78.5}  & \textbf{95.1} & \textbf{58.4}       \\
\bottomrule
\end{tabular}}
\caption{Experimental results on four datasets. $|\theta|$ is the number of parameters on the SNLI task. * means the model utilize the syntax tree of sentences. The evaluation metrics on both tasks are the accuracy. The underlining means the best result among baseline models without BERT.}
\label{tab:1}
\end{table*}

\section{Experiments}
We evaluate the proposed MS-SAN on the natural language inference (NLI) and the sentiment classification (SC) tasks. The goal of NLI is to reason the semantic relationship between a premise and a hypothesis, containing entailment, natural and contract. We experiment on the Stanford Natural Language Inference (SNLI) dataset\cite{bowman2015large} and the Multi-Genre NLI dataset (MNLI) \cite{williams2017broad}. Meanwhile, SC is a task of classifying the sentiment in sentences. We conduct experiments on the Stanford Sentiment Treebank (SST) \cite{socher2013recursive} dataset to evaluate our model in single sentence classification. We experiment SST-2 and SST-5 dataset labels with binary sentiment labels and five fine-grained labels, respectively. 

Since using pre-trained models is another feasible solution to improve the model performance in a parallel direction with the assistance of the large unsupervised corpus, such as BERT \cite{devlin2019bert}, we conduct two comparison experiments. One excludes these models in comparison and focuses on the relevant models to verify the efficiency of structural priors in resource-constrained conditions. The other is based on BERT, which replaces the embedding layer with the BERT encoder, for studying how the MS-SAN behaves when meeting BERT and verifying if the structural priors can make assist to the pre-trained models.${0,0\infty}^{lxl}$

\subsection{Experiment Setting \label{exp}} 
For both tasks, we parse the sentences by Stanford Parser \cite{chen2014fast} to calculate the dependency distance. We stack 1 layer of the MS-SAN encoder with 6 attention heads, where the mask matrices applied for each head are described in Section \ref{comb} and shown in Figure \ref{fig:2}(iii). We treat the sentence vector extracted by the MS-SAN as the classification feature for the SC task. Meanwhile, for the NLI task, we treat our MS-SAN as a sentence encoder and do the comparison with other sentence encoders. We follow the standard procedure in \cite{bowman2016fast}, and treat $concat(r_p, r_h, r_p*r_h, |r_p-r_h|)$ as the classification feature, where $r_h$ and $r_p$ are representations of the premise and the hypothesis. Finally, we feed the classification feature into a 2-layers FFN for prediction. All experiments run on a 12G NVIDIA Titan X GPU with a batch size of 32. More details are shown in Appendix.


\subsection{Experiment Results}
\subsubsection{Comparison with Traditional Encoders}
As shown in Table \ref{tab:1}, MS-SAN achieves the best performance on the two tasks. Traditional sentence encoders are mainly based on the LSTM. Among them, we select the HBMP\cite{talman2019sentence} as a representative. Comparing with HBMP which contains 3 BiLSTM layers, MS-SAN performs better with fewer parameters and higher parallelizability. Comparing with tree-based models containing Tree-CNN \cite{mou2016natural}, TreeLSTM \cite{choi2018learning,tai2015improved} and PSAN \cite{wu2018phrase}, MS-SAN models the latent hierarchical structure through the dependency distance masks without hierarchical modeling and gains a higher accuracy. Besides, DSA \cite{yoon2018dynamic} performs better than most of the baselines thanks to the dense-connected CNN encoders and the dynamic self attention for extracting feature vectors.

Among the SAN-based models, the standard Transformer performs worst. Star-Transformer \cite{guo2019star} models local dependencies through its special ring connections and achieves better performance. DiSAN \cite{shen2018disan} gains an improvement by introducing the directional information and the attentive pooling. The distance-based SAN \cite{im2017distance} integrates the word distance mask into the DiSAN and achieves a further improvement. However, the two models treat the self attention as a whole, employing two SANs with forward masks and backward masks respectively. In comparison, MS-SAN integrates the above two priors along with the dependency distance into one encoder. Therefore, MS-SAN contains fewer parameters than the above two models and achieves a better performance.

\subsubsection{Comparison with BERT Encoders}
For studying how the MS-SAN performs when meeting BERT, we implement a variant of the MS-SAN, called $\text{MS-SAN}_{\text{bert}}$. We replace the embedding layer with the BERT encoder and use the [CLS] outputs of the BERT instead of the max-pooling results in our $\text{MS-SAN}_{\text{bert}}$. For a fair comparison, we fine-tune the BERT and the $\text{MS-SAN}_{\text{bert}}$ in the same way as training the original MS-SAN described above. The experiment result is shown in Table \ref{tab:1}. As we can see, BERT performs best among the baseline models on most of the datasets except the SNLI. Thanks to the multiple structural priors, we can achieve better performance on all datasets when adding the MS-SAN on the top of BERT. Therefore, even if BERT has learnt latent sentence structures from large scale corpus, structural priors can also assist BERT on modeling sentence structures and further improve its capability of understanding sentences. However, BERT-large can learn more linguistic knowledge from larger corpus. Therefore, comparing with results at the scale of BERT-base, the improvement on the BERT-large decreases.

\subsection{Ablation Study} 
\subsubsection{Analysis of the Structural Priors}

\begin{table}
	\centering
	\begin{tabular}{lc}  
	\toprule
	    Model &Acc.(d/t)\\
	\midrule
		Standard SAN                &86.5/85.6\\   
	\midrule
        \quad + direction      &86.9/86.3\\
        \quad + word distance       &86.9/86.3\\
        \quad + dp. distance &86.8/86.3\\
    \midrule
        \quad + word \& dp. distance        &87.0/86.6\\
        \quad + direction \& word distance       &87.1/86.6\\
        \quad + direction \& dp. distance &87.0/86.7\\
	\midrule
		\quad + all mask (MS-SAN)   &87.4/87.0\\
	\bottomrule
	\end{tabular}
	\caption{Ablation study for different structure priors on the SNLI dataset. Acc.(d/t) means the accuracy on dev/test set. ``+'' denotes applying that structure prior on the SAN.}
\label{tab:3}
\end{table}

We conduct an ablation study to analyze the influence of different structure priors. The experiment result is shown in Table \ref{tab:3}. 
The standard SAN performs worst due to the lack of structural priors. 
The direction mask plays an important role in modeling sentence representations and improves the performance of the SANs thanks to the sequential order information.
Besides, no matter which distance mask is applied, both the unidirectional models and the bidirectional models perform better.
Furthermore, if applying all structural priors guided masks, our model can gain further performance improvement.
The experiment results demonstrate the fact that no matter which kind of structure prior can benefit our model on capturing structure information and dependencies between words. In addition, multiple structure priors can complement each other and bring greater benefits to our model.

\subsubsection{Analysis of the MM-MH Attention}
The core of aggregating multiple kinds of structural priors into one encoder is MM-MH attention. Thanks to the multi-mask strategy, there is no need to apply more SAN encoders to capture different priors, especially the direction masks. For verifying the effectiveness of the MM-MH attention, we compare our model with its variant $\text{MS-SAN}_{\text{sep}}$. Similar to the DiSAN \cite{shen2018disan}, it utilizes two MS-SAN encoders for modeling forward and backward information individually and concatenates the two encoding results together as the final sentence representation. 
As shown in Table \ref{tab:4}, MS-SAN extracts the sentence vector with a half dimension comparing with the $\text{MS-SAN}_{\text{sep}}$ and performs better. Besides, the results of the time-consuming experiment show the MS-SAN could be 2 times faster than the $\text{MS-SAN}_{\text{sep}}$ in the training process. Thanks to the multi-mask strategy, MS-SAN could capture the information from two directions in one encoder and greatly reduce the training time and memory consumption. Furthermore, the ``all-in-one'' model can better integrate information from different priors, thereby generating more proper sentence representations.

\begin{table}
	\centering
	\begin{tabular}{lcccc}  
		\toprule
		                        &$|\theta|$ & Dim.  &Acc.(d/t)       &Time\\
		\midrule
		MS-SAN                  &1.8m       &600    &87.4/87.0  &2ms\\ 
		$\text{MS-SAN}_{\text{sep}}$ &4.0m       &1200   &86.9/86.6  &4ms\\
		\bottomrule
	\end{tabular}

	\caption{Comparison experiments between MS-SAN and $\text{MS-SAN}_{\text{sep}}$ on SNLI dataset. $|\theta|$ is the number of model parameters. Time is calculated as millisecond per batch during training (batch size is 32). Acc.(d/t) means the accuracy on SNLI dev/test set.}
	\label{tab:4}
\end{table}

\subsubsection{Comparison with the Relative Position Embedding}
\begin{table}
	\centering
	\begin{tabular}{p{0.73\columnwidth}c}  
	\toprule
	    Model & Acc.(d/t)\\
	\midrule
		Standard SAN                &86.5/85.6\\   
	\midrule
        \quad + rel pos \cite{shaw2018self}     &86.7/86.0\\
        \quad + xl pos \cite{dai2019transformer}        &87.1/86.3\\
        \quad + distance masks (word \& dp.) & 87.0/86.6\\
    \midrule
        \quad + direction \& rel pos       &87.0/86.2\\
        \quad + direction \& xl pos   &87.2/86.6\\
	\midrule
		\quad + direction \& distance masks (MS-SAN)   &87.4/87.0\\
	\bottomrule
	\end{tabular}
	\caption{Comparison experiment between SANs with structural priors and SANs with relative position embeddings. Acc.(d/t) means the accuracy on dev/test set.}
\label{tab:5}
\end{table}
The relative position encoding (RPE) is another flexible method to model the sentence latent structure. Comparing with the absolute position encoding method, the RPE transforms the relative position between words into vector representations and calculate an extra position-aware attentive score through them, which help the SANs capturing the relative position relationship between words. In comparison with the RPE which can learn more flexible position relationships, the mask matrices extracted from structural priors directly provide the position-aware attention bias to the SANs. Therefore, the mask mechanism can also be seen as a specific kind of the RPE, which is more simple and intuitive. 

In this subsection, we replace the structural mask matrices with two commonly used RPE, called rel-pos introduced in \cite{shaw2018self} and xl-pos from Transformer-XL \cite{dai2019transformer}, and conduct a comparison experiment. Besides, we apply the direction mask on the two RPE-based SANs to do further research. The experimental results shown in Table \ref{tab:5} demonstrate that both the RPE and the structural priors can assist SANs on modeling sentence structures. Comparing with using a single RPE, multiple kinds of distance masks can complement each other and improve the performance. Meanwhile, direction masks can also provide stronger directional information for both the two RPEs. Therefore, we can treat the RPE as one kind of learnable structure priors and merge them with other structure priors through our model.
\begin{figure*}[!t]
	\centering
	\includegraphics[scale=0.18]{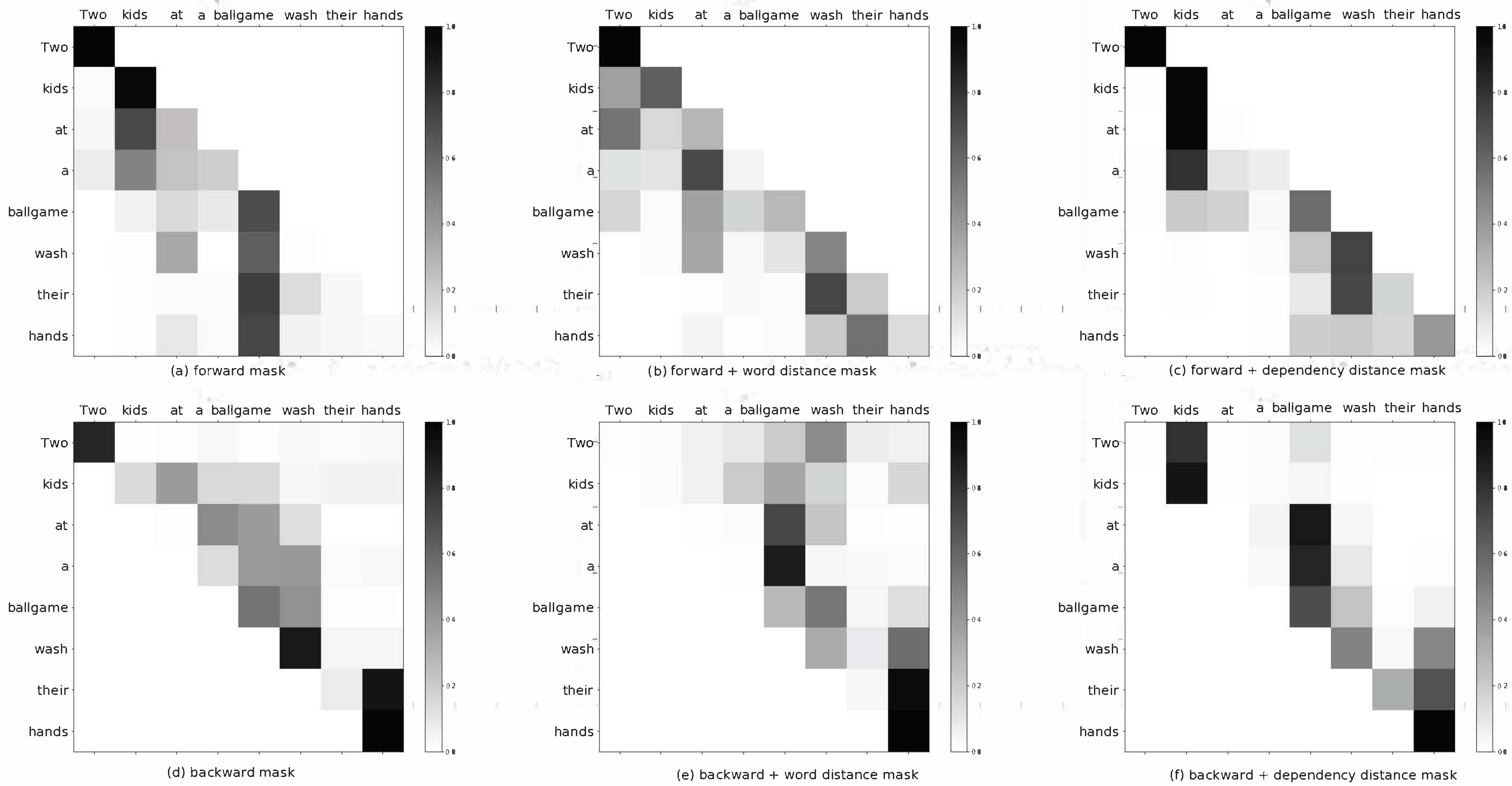}
	\caption{Heat maps of the multiple structural priors guided multi-head attention weights, with the example sentence of ``Two kids at a ballgame wash their hands''.}
	\label{fig:4}
\end{figure*}

\subsection{Case Study}
In order to further analyse how the structural priors work, we visualize the attention distribution of each head in the MS-SAN. We take the sentence ``Two kids at a ballgame wash their hands'' as an example, which dependency syntax tree is shown in Figure \ref{fig:3} and mask matrices is shown in Figure \ref{fig:2}. 

As shown in Figure \ref{fig:4}, each head has a different focus. Under the direction masks, one attention head can only focus on the words before or after the reference word. The normal head focus more on the keywords such as ``kids, ballgame, wash and hands'' and the word distance masked heads focus more on the neighboring words. Meanwhile the dependency distance masked heads focus more on the dependent relative words on the dependency syntax tree. For instance, all words in ``at a ballgame'' pay more attention on the word ``ballgame'' since the three words compose a sub-tree in the dependency parsing results. Besides, comparing with the backward word distance masked head, ``wash'' pay less attention on ``their'' because of the longer distance between them on the dependency syntax tree. This demonstrates that the dependency distance can strengthen the ability of modeling latent hierarchical structure of sentences and can capture more precise dependency of words. Furthermore, different structural priors have different effects on the model, allowing the model to understand texts from more perspectives, thereby further improve the model's modeling capabilities.

\section{Related Work}
Recently, self-attention mechanism has achieved great success.  \citeauthor{vaswani2017attention} \shortcite{vaswani2017attention} proposed the Transformer architecture and achieved great improvements on the machine translation task. Then, \citeauthor{yu2018qanet} \shortcite{yu2018qanet} proposed the QA-Net for the reading comprehension task. In addition, the language models based on transformers pre-trained on large corpus bring a huge improvement on many NLP tasks \cite{devlin2019bert,yang2019xlnet}. 

However, standard SANs is limited in modeling sentence structures, including the sequential and the hierarchical structure. For the former, previous studies follow the idea of BiLSTMs and CNNs, introducing similar information into SANs. \citeauthor{shen2018disan} \shortcite{shen2018disan} proposed the DiSAN, introducing the direction information into SANs. After that, \citeauthor{im2017distance} \shortcite{im2017distance} introduced the word distance on the basis of the DiSAN in order to capture more local dependencies. For the latter, traditional methods are mostly based on dependency syntax trees or constituency syntax trees \cite{li2015tree,tai2015improved,mou2016natural}, which is hard to parallelize with a huge training cost. However, SAN-based models can leverage these limitations with the assistance of the self-attention. 
Both PSAN proposed by \citeauthor{wu2018phrase} \shortcite{wu2018phrase} and Tree-Transformer proposed by \citeauthor{wang2019tree} \shortcite{wang2019tree} utilized constituency trees to model sentences in phrase level. \citeauthor{wang2019self} \shortcite{wang2019self} improved the position encoding through a relative structural position extracted from dependency trees for modeling latent hierarchical structure. Our model merges both the sequential and hierarchical structure information through the multiple structural priors guided self attention.

\section{Conclusion}
We propose the Multiple Structural Priors guided Self Attention Network (MS-SAN), which utilize multiple types of structural priors to model texts. By applying the multi-mask strategy, we can take full use of the multi-head attention to capture the information of texts at multiple aspects. We introduce two categories of structural priors into the MS-SAN, containing the sequential order and the relative position of words. Thanks to these priors, MS-SAN gains better understanding of the texts with a stronger ability of modeling latent sentence structure. Experiments on the NLI and SC tasks also demonstrate the effectiveness of our model. In the future, we will try to combine relational position embeddings with the structural priors together for modeling sentence structures better. Besides, we will try to introduce structural priors into pre-trained models for model reduction.

\bibliography{aaai21}
\end{document}